\RequirePackage{algorithm}
\RequirePackage{algorithmic}

\documentclass{article}

\PassOptionsToPackage{square, numbers}{natbib}

\usepackage[preprint]{neurips_2021}

\usepackage[utf8]{inputenc} 
\usepackage[T1]{fontenc}    
\usepackage{hyperref}       
\usepackage{url}            
\usepackage{booktabs}       
\usepackage{amsfonts}       
\usepackage{nicefrac}       
\usepackage{microtype}      
\usepackage{xcolor}         

\usepackage{caption}
\usepackage{amssymb}
\usepackage{mathtools}
\usepackage{multirow}
\usepackage{graphicx}
\usepackage{wrapfig}
\usepackage{pifont}
\usepackage{verbatim}

\newcommand{\cmark}{\ding{51}}

\title{Meta-learning Amidst Heterogeneity and Ambiguity}

\author{%
  Kyeongryeol Go \\
  Graduate School of AI \\
  KAIST \\
  Daejeon, South Korea \\
  \texttt{kyeongryeol.go@kaist.ac.kr} \\
  \And
  Seyoung Yun \\
  Graduate School of AI \\
  KAIST \\
  Daejeon, South Korea \\
  \texttt{yunseyoung@kaist.ac.kr} \\
}

\begin{document}

\maketitle

\begin{abstract}
  Meta-learning aims to learn a model that can handle multiple tasks generated from an unknown but shared distribution. However, typical meta-learning algorithms have assumed the tasks to be similar such that a single meta-learner is sufficient to aggregate the variations in all aspects. In addition, there has been less consideration on uncertainty when limited information is given as context. In this paper, we devise a novel meta-learning framework, called Meta-learning Amidst Heterogeneity and Ambiguity (MAHA), that outperforms previous works in terms of prediction based on its ability on task identification. By extensively conducting several experiments in regression and classification, we demonstrate the validity of our model, which turns out to be robust to both task heterogeneity and ambiguity.
\end{abstract}

\section{Introduction}

Although deep learning models have shown remarkable performance in various domains, they have consistently been criticized because of their sensitivity to the amount of data \citep{chen2014big, najafabadi2015deep, cho2015much, sun2017revisiting, hestness2017deep}. Despite all available public data, the data scarcity issue is still not negligible. In many cases, the actual data that is worth analyzing is quite limited for many different reasons, for example, concerns about data privacy \citep{liu2020machine} and noisy data with anomalies \citep{sanders2017garbage}. Along with transfer learning, few-shot learning, and multi-task learning, meta-learning has recently been highlighted as a way to overcome this deficiency with its adaptive behavior using a few data points \citep{vanschoren2018meta, hospedales2020meta}. 

Meta-learning aims to handle multiple tasks by efficiently organizing the acquired knowledge. However, typical algorithms have been assessed based on a solid assumption which lacks the representative potential in real-world scenarios. Among many tackles \citep{triantafillou2019meta, lee2019learning}, we mainly focus on the following two assumptions. First of all, the tasks are regarded to be similar such that a single meta-learner is sufficient to aggregate the variations in all aspects. It implies that there has been little effort to compactly abstract notions within heterogeneity, one of the essential factors characterizing human intelligence, which is advantageous in decision-making to query the associated information to solve the problem. In addition, there has been less consideration on uncertainty for identifying particular task with a few data points. It is therefore not easy to analyze or transfer the acquired knowledge of the model, which is critical in the growing AI industries, such as a medical diagnosis \citep{ahmad2018interpretable, challen2019artificial, vellido2019importance} and autonomous vehicles \citep{kim2017interpretable, shafaei2018uncertainty, chen2020interpretable}, because a certain level of interpretability is required for greater safety.

In this respect, we hypothesize that a disentanglement in task representation is advantageous, which frequently appears in studies to analyze the inherent factors of variation within the dataset. This is to \textit{i) uncover the distinctive properties as a tool for interpretability} and to \textit{ii) explicitly separate the dataset into several clusters, which would have been detrimental when trained altogether}. However, as a trade-off for interpretability, the overconfident nature of deep learning may strictly assign the tasks into certain clusters without considering ambiguity, which requires an additional treatment to cope with the anomalies.


\begin{wrapfigure}[13]{r}{0.4\textwidth}
    \centering
    \includegraphics[width=0.4\textwidth]{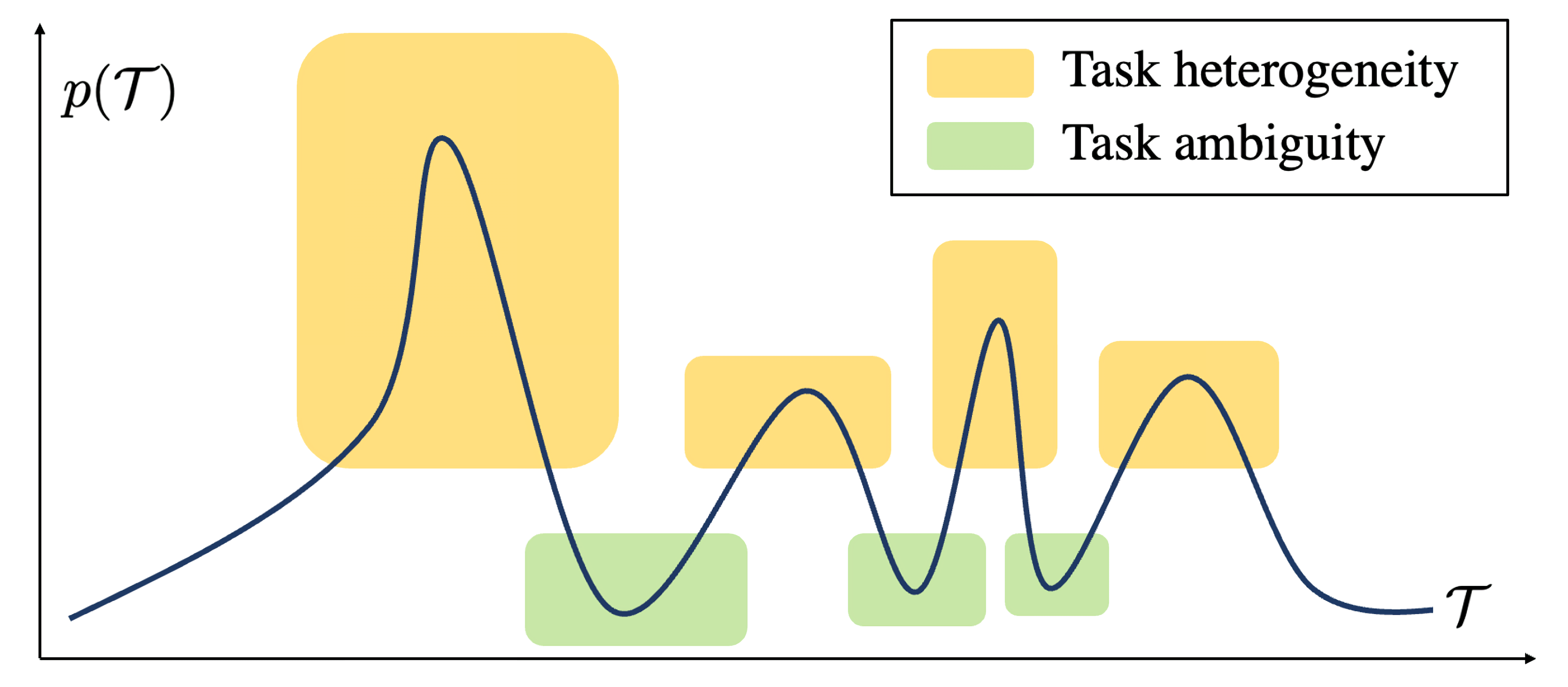}
    \caption{Heterogeneity and ambiguity occurred in task distribution. Those are not independent concepts, but the ambiguity naturally comes after the heterogeneity.} 
    \label{task distribution}
\end{wrapfigure}

To this end, we propose a new meta-learning framework, Meta-learning Amidst Heterogeneity and Ambiguity (MAHA), that performs robustly on the following two huddles. \textbf{Task heterogeneity}: there is no clear discrimination between the tasks that are sampled from the faraway modes of task distribution \citep{vuorio2018toward, yao2019hierarchically, yao2020automated}. \textbf{Task ambiguity}: too few data points are given to infer the task identity \citep{finn2018probabilistic, rusu2018meta}. Specifically, we devise a pre-task built upon the neural processes \citep{garnelo2018conditional, garnelo2018neural, kim2019attentive} to obtain well-clustered and interpretable representation. Then, an agglomerative clustering is applied to the representation without any external knowledge such as the number of clusters and separately train a different model for each cluster. Please refer to Figure~\ref{model diagram} for the overall training process of MAHA.

To summarize, the main contributions of this paper are the following 4-folds:
\begin{itemize}
    \item We propose a simple yet powerful architecture design for the neural processes to better leverage the latent variables and be applicable in classification. (See Section~\ref{encoder-decoder pipeline})
    \item We resolve the information asymmetry in the neural processes and construct well-clustered and interpretable representations. (See Section~\ref{method:disentanglement})
    \item We validate MAHA through both regression and classification, by which the experimental results demonstrate its ability to cope with the heterogeneity and ambiguity. (See Section~\ref{experiment})
    \item We devise an additional regularization term for the low-\textit{shot} regime that distills an obtainable knowledge from relatively various training samples and variations. (See Appendix B)
\end{itemize}

\section{Related work}

Gradient-based meta-learning, represented by MAML \citep{finn2017model}, aims to learn the prior parameters that can quickly adapt to certain tasks through several gradient steps. It consists of the inner-loop for the task adaptation and the outer-loop for the meta-update over tasks. Many variants have emerged to balance generalization and customization in a task-adaptive manner. To begin with a generalization perspective, \citep{finn2018probabilistic, kim2018bayesian} suggested probabilistic extensions through the hierarchical Bayesian model and Stein variational gradient descent (SVGD) \citep{liu2016stein}. In addition, \cite{rusu2018meta} conducted the inner-loop on the low-dimensional latent embedding space, and \citep{yin2019meta} proposed the meta-regularization that was built on information theory. From a customization perspective, \citep{lee2018gradient} divided the parameters into two categories, one of which is shared across tasks, and the other can be modulated task-specifically. \citep{zintgraf2019fast} was informed by the layer-wise adaptive units, and \citep{vuorio2018toward, yao2019hierarchically, yao2020automated, yao2020online} considered the auxiliary networks that modulate the initial parameter before the inner-loop.

The family of neural processes, also known as contextual meta-learning, is devised to imitate the flexibility of the Gaussian Process \citep{rasmussen2003gaussian} while resolving the scalability issue. Rather than explicitly modeling the kernel to conduct the Bayesian inference like \citep{wilson2016deep}, it learns an implicit kernel directly from data which overcomes the design restrictions. Task-specific information is extracted from the subset of data through an encoder, which is then aggregated for utilization in the decoder to predict the corresponding outputs of the remaining data. Starting from the conditional neural process (CNP) \citep{garnelo2018conditional}, which was built solely on a deterministic path, the neural process (NP) \citep{garnelo2018neural} applies the addition of a stochastic path. The attentive neural process (ANP) \citep{kim2019attentive} further applies an attention mechanism to resolve the underfitting issue in NP by enlarging the locally adaptive behavior. More complex modules, such as a graph structure \citep{louizos2019functional} and recurrent neural network \citep{kumar2018consistent, singh2019sequential}, were further considered to capture the dependencies on latent variables and the complex temporal dynamics.

However, many problems remain unsolved. Firstly, the neural processes yet rely on a complex feature extractor to enable task-specific modulation, which requires various regularization techniques with additional hyperparameters \citep{requeima2019fast}. Furthermore, whereas the neural processes are able to obtain an explicit task representation, the existing approaches have investigated little regarding interpretability. Finally, the performance analysis has been mainly focused on regression \citep{le2018empirical, kim2019attentive, singh2019sequential, sureshimproved, gordon2020convolutional}, and some are not even directly applicable for classification \citep{lee2020bootstrapping}.


\section{Problem setting}

Let $C=\{C_x, C_y\}$ be the context set, and let $T=\{T_x, T_y\}$ be the target set, where both $C$ and $T$ are sampled from the same task $\mathcal{T} \sim p(\mathcal{T})$. A common goal in meta-learning is to devise an algorithm for the model $f(\cdot)$ that appropriately uses the model parameter $\theta$ to obtain the task-specific parameter $\phi$ according to the input-output pairs in $C$ such that when $T_x$ is given, $T_y$ can be accurately estimated with high confidence. 
For example, in MAML \cite{finn2017model}, a task-specific parameter can be computed by using a gradient step $\phi = \theta - \alpha \cdot \nabla_{\theta} L(f(C_x;\theta), C_y)$. On the other hand, in CNP \citep{garnelo2018conditional}, $\theta$ and $\phi$ no longer share the same parameter space. Here, the model parameter is divided into an encoder and a decoder part $\theta=\{\theta_\text{enc}, \theta_\text{dec}\}$, and the task-specific parameter can be computed by the encoder output $\phi = f_\text{enc}(C;\theta_\text{enc})$. Hereafter, we omit $\theta$ for brevity.

For model training, $\phi$ is iteratively updated using \textit{batch}s. Here, each \textit{batch} is constructed through multiple tasks that are characterized by \textit{way} and \textit{shot}. If there are N classes, each of which contains K input-output pairs, we call it an N-\textit{way} K-\textit{shot} problem. The class labels are shuffled in classification whenever a task instance is created, which encourages a meta-learning algorithm to learn how to classify images even when the configuration of unseen classes occurs.

\section{Preliminary : (Attentive) Neural Process}
\label{(A)NP}


\begin{wrapfigure}[21]{r}{0.35\textwidth}
    \centering
    \includegraphics[width=0.34\textwidth]{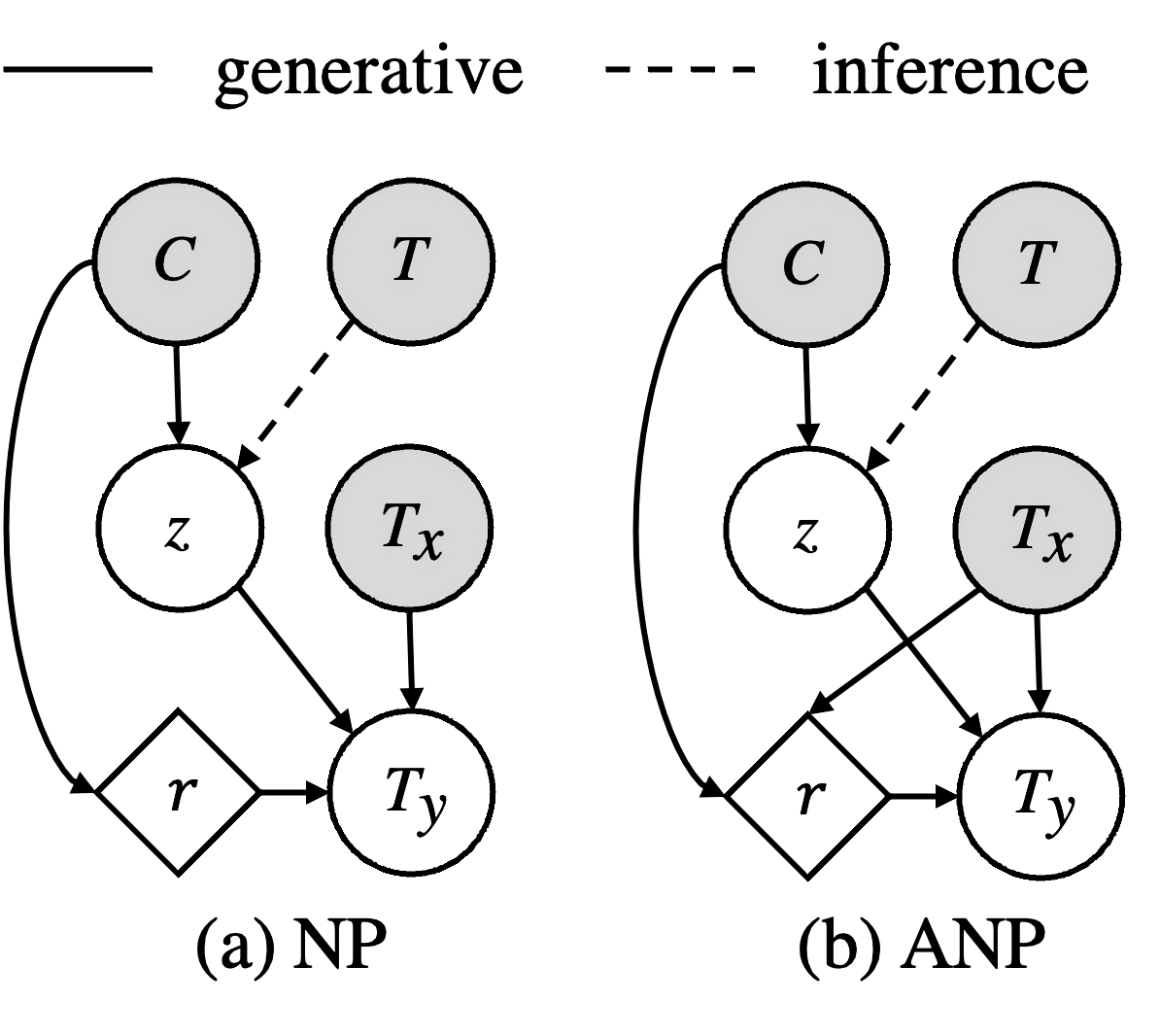}
    \caption{Graphical model of the related baselines. Circles denote random variables, whereas diamonds denote deterministic variables. Shaded variables are observed during the test phase, and every in-between edge is implemented as a neural network.} 
    \label{graphical_model}
\end{wrapfigure}

In Figure~\ref{graphical_model}, we summarize how a basic family of neural processes has evolved in terms of the graphical model. The encoder comprises a deterministic path and stochastic path computing the task-specific parameter $\phi=\{r, z\}$ of the variational distributions which we denote by $q(r|\{X, Y\})=\mathcal{N}(r, 0)$ and $q(z|\{X, Y\})=\mathcal{N}(\mu_z, 0.1 + 0.9 \cdot \text{sigmoid}(\omega_z))$.\footnote{Note that $r$ is deterministic with zero variance.} Here, $\{X, Y\}$ indicates a set of input-output pairs and a reparameterization trick is applied at the end of the stochastic path for differentiable non-centered parameterization \citep{kingma2014efficient}.

For both paths, NP is constructed by:
\begin{align*}
    & r=\text{MeanPool}_\textit{shot}(\text{rFF}(\{X, Y\})) \\ 
    & [\mu_z, \omega_z]=\text{MeanPool}_\textit{shot}(\text{rFF}(\{X, Y\}))
\end{align*}
where MeanPool($\cdot$) is a mean-pooling operation along the subscripted dimension, rFF($\cdot$) can be any row-wise feedforward layer, such as Multi-Layer Perceptron (MLP), and $[\cdot]$ denotes the concatenation. On the other hand, ANP exploits the multi-head attention, connecting $T_x$ to $r$ in graphical model, and self-attention, both of which are proposed in \citep{vaswani2017attention}.
As in NP, the value of $z$ is same for every \textit{shot} of $T_x$, however, based on the attention score with each element of $X$, $r$ is now computed in \textit{shot}-dependent manner. 

Then, conditioned on the encoder outputs, $r$ and $z$, with the target input $T_x$, the decoder computes the parameters of predictive distribution on the target output $T_y$:
\begin{equation}
    \label{(A)NP decoder}
    [\mu_{T_y}, \omega_{T_y}] = \text{rFF}\left([T_x, r, z]\right)
\end{equation}
where the predictive distribution is expressed as $p(T_y|T_x, r, z)=\mathcal{N}(\mu_{T_y}, 0.1 + 0.9 \cdot \text{softplus}(\omega_{T_y}))$. Eventually, relying on the variational inference, one can obtain the loss function which approximates the negative ELBO by replacing an intractable $p(z|C)$ with the variational distribution $q(z|C)$ following \citep{garnelo2018neural}:
\begin{equation}
    \label{(A)NP loss}
    \mathcal{L}_{(A)NP} = - \mathbb{E}_{q(r|C)q(z|T)} \left[ \log{p(T_y|T_x, r, z)} \right] + \beta_{1} KL \left( q(z|T) \| q(z|C) \right)
\end{equation}

As a result, based on the Kolmogorov extension and de-Finetti theorems, the neural processes become a stochastic process that satisfies the exchangeability and consistency \citep{garnelo2018neural}. However, when trained using the deterministic path, the neural processes with latent variables is empirically shown to have difficulty capturing the variability of the stochastic process \citep{le2018empirical}, of which causes are investigated and resolved in Section~\ref{method:disentanglement}. 




\section{Meta-learning Amidst Heterogeneity and Ambiguity}
\label{methodology}


This section describes our algorithm MAHA whose primary focus is to devise a pre-task to cope with task heterogeneity and ambiguity in meta-learning. We first introduce an encoder-decoder pipeline of MAHA, namely FELD, of which effects are examined by substituting the correspondent within NP in Section~\ref{experiment}. Then, a dimension-wise pooling and an auto-encoding structure are proposed to obtain well-clustered and interpretable representation. Finally, the training process of MAHA is described, which applies to both regression and classification. 

\subsection{Encoder-decoder pipeline}
\label{encoder-decoder pipeline}

\begin{figure}[h]
    \centering
    \includegraphics[width=0.8\textwidth]{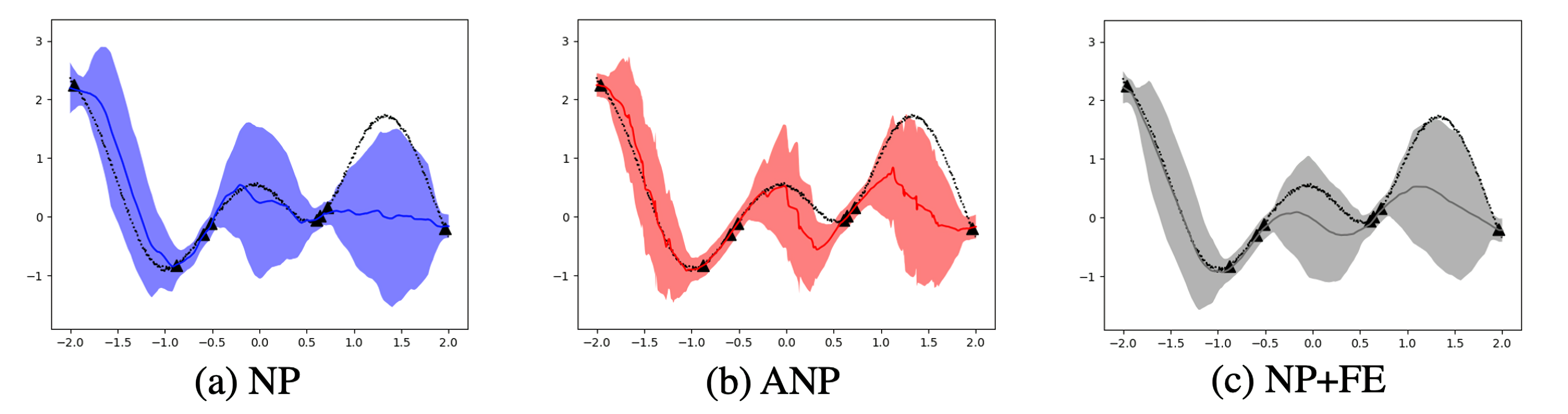}
    \caption{Qualitative comparison between NP, ANP, and NP with the flexible encoder (NP+FE) on functions generated from Gaussian Process. The shaded areas correspond to the $\pm$2 standard deviations. Prediction of ANP turns out to be wiggly, while NP and NP+FE are relatively smooth following Occam's razor. Note that quantitative comparison can be looked up in Table~\ref{gp regression MSE}.} 
    \label{gp regression}
\end{figure}

\paragraph{Flexible Encoder} Although the attention mechanism proposed in ANP was a key to resolve the underfitting in NP, there is less incentive for $r$ to focus on task identity that is shared across \textit{shot}s. As a result, in Figure~\ref{gp regression}, ANP appears to strongly fit the given input-output pairs, which leads to a wiggly prediction. Particularly within task heterogeneity and ambiguity where the prediction space is prone to be highly variable, the wiggly prediction of ANP leads to a poor generalization performance (See Figure~\ref{regression performance}). Therefore, the graphical model of NP is rather considered in MAHA since its latent variables are \textit{shot}-independent. Then, based on analysis in \citep{cremer2018inference}, the problematic underfitting is dealt with by substituting the encoder with the flexible and permutation-invariant Set Transformer (ST) \citep{lee2019set}. Note that the Set Transformer can incorporate the rFF($\cdot$) and $\text{MeanPool}_\textit{shot}(\cdot)$ in the encoder of NP. See Appendix A for a more detailed explanation about the modules in Set Transformer.

\paragraph{Linear Decoder} We avoid using a complex decoder such as \citep{oord2016pixel} and apply feature-wise linear modulation to the target input $T_x$. Inspired by \citep{zhao2017learning}, we composite the latent variables using a skip connection. Among the many normalization techniques, a layer normalization \citep{ba2016layer} is applied since the statistic is computed independently for each \textit{batch} instance such that only $z$ can capture the heterogeneity in accordance with the pooling proposed in Section~\ref{method:disentanglement}.
\begin{align}
    \label{MAHA linear}
    [\mu_{T_y}, \omega_{T_y}] \quad \text{or} \quad logit = \text{g}(T_x) \cdot W \quad \text{where} \quad W = \text{w}(r,z) = \text{LN}(r + \text{rFF}(z))^\text{T}
\end{align}
Here, g($\cdot$) implies any feature extractor, LN($\cdot$) indicates a layer normalization, and the transpose operation $\text{T}$ permutes the last two dimensions of the tensor. It is aligned with the previous approaches \cite{bowman2015generating, semeniuta2017hybrid, yang2017improved, he2019lagging} which weaken the decoder to allow the latent variables to be appropriately leveraged. Also, it relates to studies on few-shot classification \citep{gordon2018meta, requeima2019fast} where each column of $W$ is computed by \textit{shot}s within the same \textit{way}. However, when accompanied with the pooling in Section~\ref{method:disentanglement}, the columns are no more independent by one another and share information across \textit{way}.

\begin{figure}[h]
    \centering
    \includegraphics[width=0.9\textwidth]{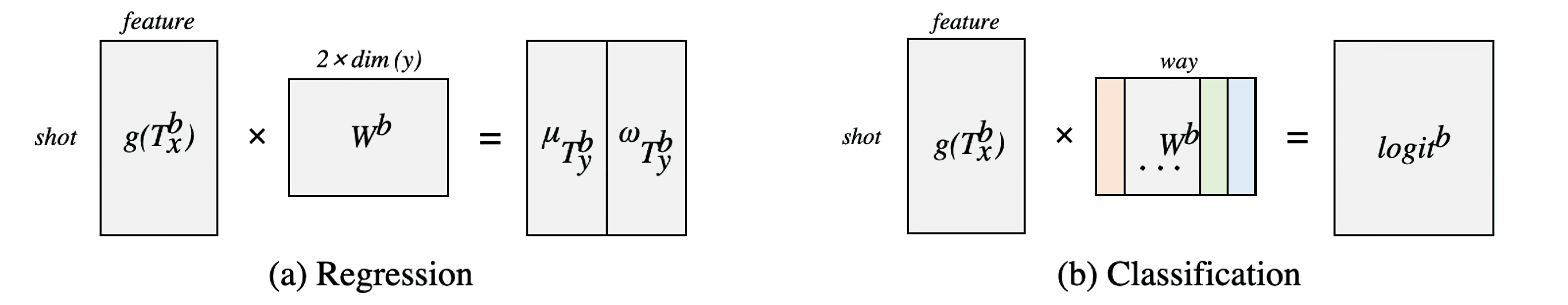}
    \caption{Prediction on output distribution. Superscript $b$ indicates the $b$-th batch instance.} 
    \label{prediction}
\end{figure}

\subsection{Inducing disentanglement on $z$}
\label{method:disentanglement}

\begin{wrapfigure}[14]{r}{0.34\textwidth}
    \centering
    \includegraphics[width=0.34\textwidth]{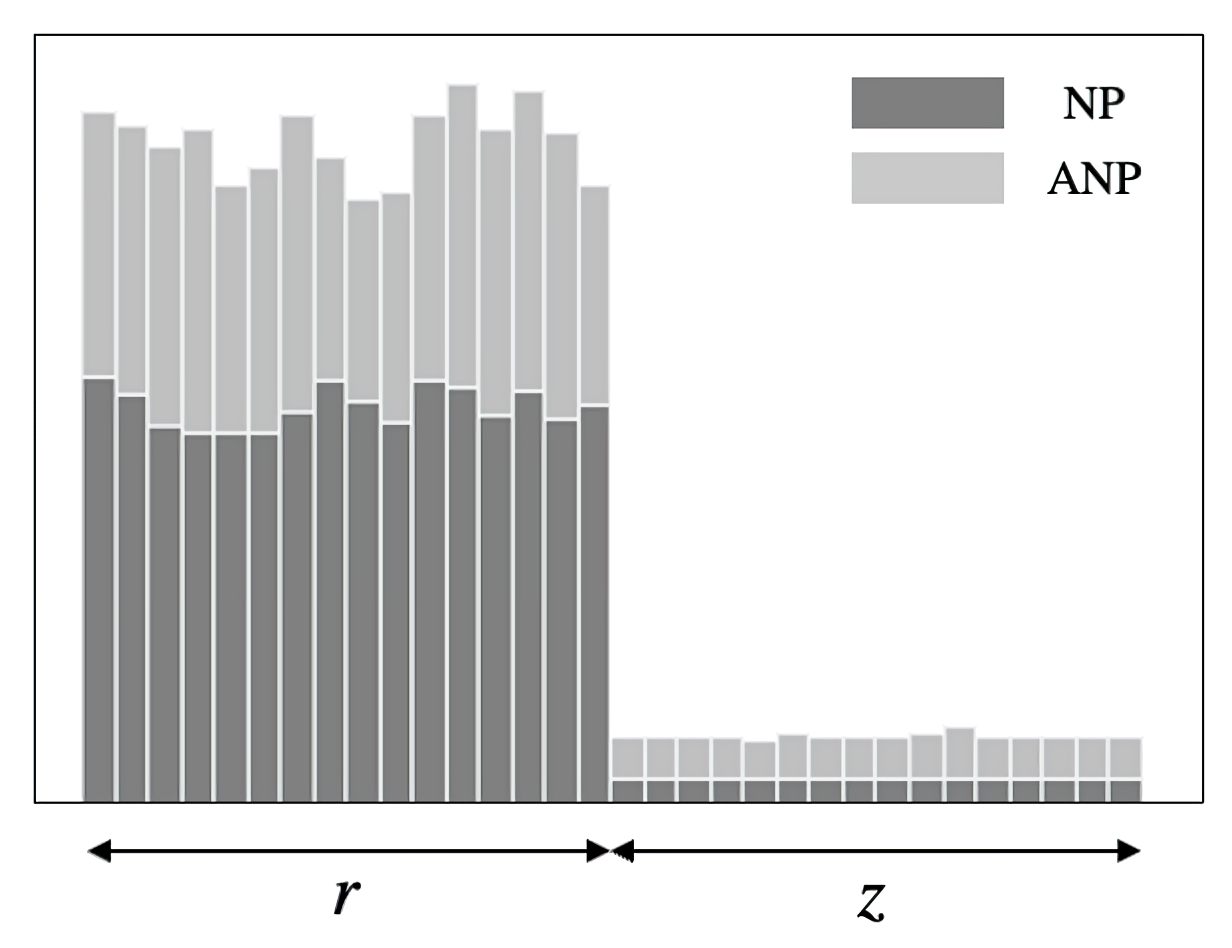}
    \caption{Stacked bar plot for the weight norm of the decoding layer}
    \label{weight norm}
\end{wrapfigure}

\begin{figure}[t]
    \centering
    \includegraphics[width=0.95\textwidth]{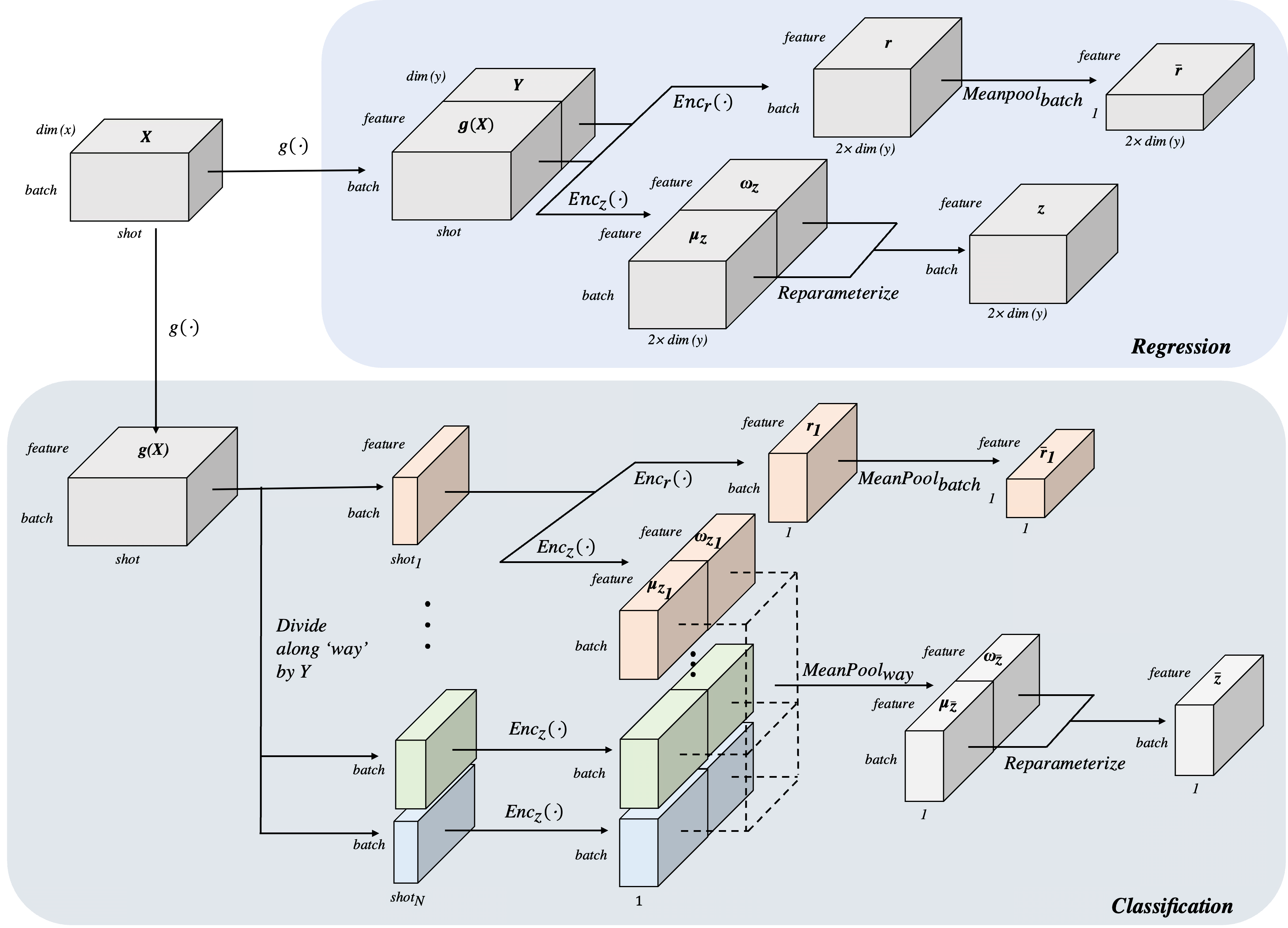}
    \caption{Computational diagram for $\bar{r}$ and $\bar{z}$. For visual comfort, every block of the encoder outputs in regression is reshaped from $[\textit{batch}, \: 1, \: 2\times \textit{dim(y)} \times \textit{feature}]$ into $[\textit{batch}, \: 2\times \textit{dim(y)}, \: \textit{feature}]$. In classification, \textit{shot} dimension is divided along \textit{way} with subscript $\{1, ..., N\}$ and $\bar{r}=[\bar{r}_1, ..., \bar{r}_N]$.}
    \label{architecture}
\end{figure}

For NP and ANP trained on functions generated from GP, we illustrate the weight norm of the decoding layer right behind the latent variables in Figure~\ref{weight norm}. The sparsely-coded decoder implies the redundancy of the stochastic path due to the component collapsing behavior referred to in \citep{nalisnick2016stick, joo2020dirichlet}. This phenomenon can be explained by the information preference problem \citep{chen2016variational, zhao2017towards} where the information flow is concentrated on the deterministic path with the tendency to ignore the stochastic path. 

In order to handle the information asymmetry, several solutions were proposed in studies on the generative model, such as the KL annealing scheduler \citep{bowman2015generating, fu2019cyclical} and expressive posterior approximation \citep{rezende2015variational, kingma2016improving}, but these are generally not robust to changes in model architecture. Instead, we propose a simple method to avoid redundancy of the stochastic path by encouraging it to acquire multi-modality within heterogeneity and ambiguity.

\paragraph{Dimension-wise pooling} 

We explicitly capture the distinct variations within the information flow by pooling each path across different dimensions, \textit{batch} for $r$ and \textit{way} for $z$:
\begin{equation}
    \label{MAHA encoder}
    \bar{r} = \text{MeanPool}_\textit{batch}(r) \quad \text{and} \quad [\mu_{\bar{z}}, \omega_{\bar{z}}] = \text{MeanPool}_\textit{way}([\mu_z, \omega_z])
\end{equation}
Then, the deterministic representation $\bar{r}$ becomes identical not only across \textit{shot}, but also across \textit{batch}. Then, whenever it is insufficient to handle all variations across tasks within the same \textit{batch} i.e., facing task heterogeneity, the model should resort to the stochastic representation $\bar{z}$ since the deterministic representation only captures the average properties. On the other hand, the stochastic representation $\bar{z}$ allows the different \textit{way} to share information and becomes class-invariant. We illustrate how the latent variables $\bar{r}$ and $\bar{z}$ are computed in Figure~\ref{architecture}. Note that the value of \textit{way} is set to 1 in regression such that pooling on $z$ is negligible. 


\paragraph{Auto-encoding structure} 

Empirically, we observe that the KL collapse \citep{bowman2015generating, alemi2016deep, sonderby2016ladder, zhao2017towards} does not occur whenever the pooling operations is used (see Appendix D). This implies that the posterior $q(\bar{z}|T)$ does not simply converge to the approximate prior $q(\bar{z}|C)$ so that the decoder gets dependent on the stochastic path. However, there is still an incentive for $\bar{r}$ to be underutilized during the decoding because it is inferred by small $C$ not large $T$ \citep{hewitt2018variational} and neural networks exploiting set representation is known to poorly perform in low-\textit{shot} regime \citep{edwards2016towards, zaheer2017deep} i.e., facing task ambiguity.

Thereby, we resort to the conditional auto-encoding structure \citep{sohn2015learning} on top of the dimension-wise pooling to cope with the lack of training samples. As a result, the following loss function is derived which differs from Equation~\ref{(A)NP loss} on \textit{i) whether the pooling operations are used or not} and \textit{ii) which set is used to compute the deterministic representation}, each of which is the result of the dimension-wise pooling and the auto-encoding structure:
\begin{equation}
    \label{pretraining loss}
    \mathcal{L}_{pre} = - \mathbb{E}_{q(\bar{r}|T)q(\bar{z}|T)} \left[ \log{p(T_y|T_x, \bar{r}, \bar{z})} \right] + \beta_{2} KL \left( q(\bar{z}|T) \| q(\bar{z}|C) \right)
\end{equation}

\subsection{Training process}
\label{training process}

\begin{figure}[t]
    \centering
    \includegraphics[width=\textwidth]{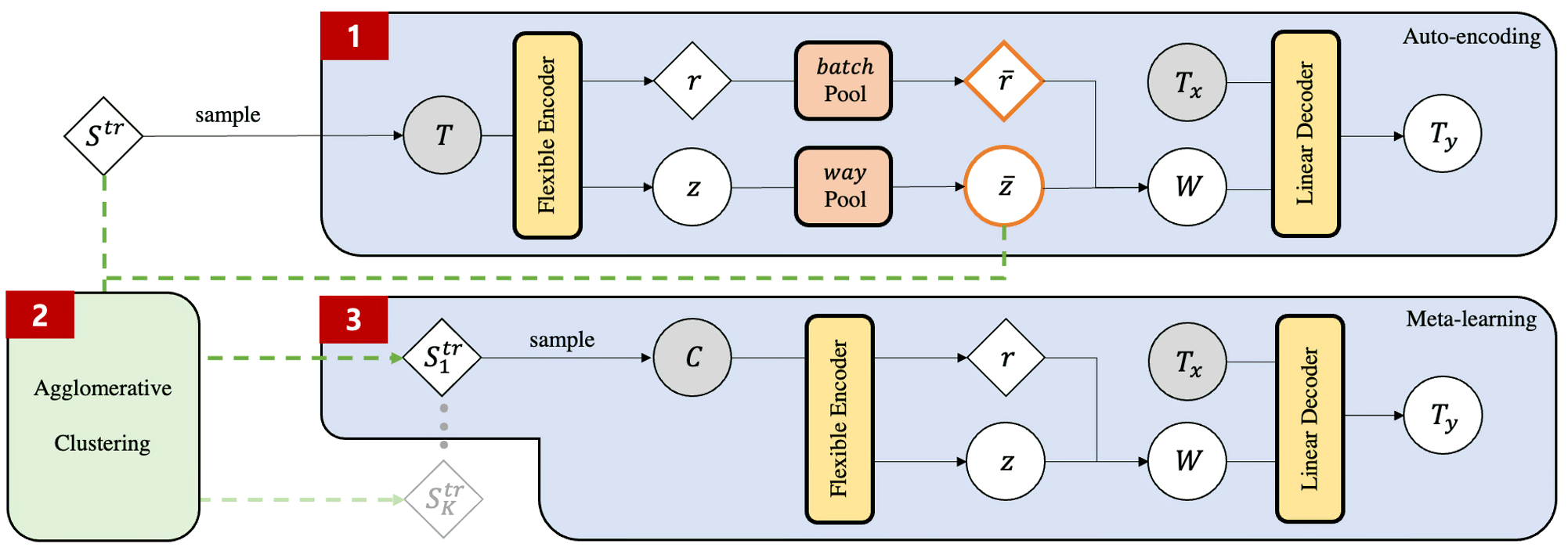}
    \caption{MAHA. $K$ is the number of estimated clusters such that the meta-train set $S^{tr}=\bigcup_{k=1}^{K} S^{tr}_{k}$.}
    \label{model diagram}
\end{figure}

See Figure~\ref{model diagram}. Initially, the dimension-wise pooling and the auto-encoding structure proposed in Section~\ref{method:disentanglement} are used along with FELD to minimize the loss function in Equation~\ref{pretraining loss}. Next, an agglomerative clustering is applied to the disentangled representation from the stochastic path to estimate the number of clusters with the highest purity value.\footnote{For a homogeneous dataset, a single cluster is available such that the previous steps can be omitted.} Finally, for each cluster, separate FELD is trained from the beginning by Equation~\ref{(A)NP loss} where the tasks are no longer uniformly sampled but statistically skewed based on the ratio of heterogeneous tasks within the cluster. According to the Euclidean distance to the cluster centers, FELD in correspondence to the closest cluster is exploited for evaluation.

\section{Experiment}
\label{experiment}

We first experiment on frequently appearing benchmark datasets in meta-learning and investigate the role of the encoder-decoder pipeline (FELD) by gradually adjusting NP. Those datasets are generally regarded to be homogeneous such that the MAHA is equivalent to FELD when assuming a single cluster as noted in Section~\ref{training process}. After that, MAHA is evaluated on heterogeneous datasets following the experimental setting of \citep{yao2019hierarchically} with the dimension-wise pooling and the auto-encoding structure in Section~\ref{method:disentanglement}, of which roles are examined in both quantitative and qualitative manner. Please refer to Appendix C for details about the data-split, architecture design, and the hyperparameter search.

Overall, we are to answer the following three questions: 
\begin{itemize}
    \item Does MAHA outperform the previous baselines in terms of prediction? (See Table~\ref{gp regression MSE} to \ref{multi-dataset accuracy})
    \item What are the benefits of using the flexible encoder and the linear decoder? (See Section~\ref{homogeneous dataset})
    \item How does the dimension-wise pooling and the auto-encoding structure contribute to obtaining well-clustered representation within heterogeneity? (See section~\ref{heterogeneous dataset})
\end{itemize} 


\subsection{Homogeneous dataset} 
\label{homogeneous dataset}

\begin{wraptable}[11]{r}{0.35\textwidth}
    \caption{MSE on Gaussian Process}
    \label{gp regression MSE}
    \begin{center}
    \begin{scriptsize}
    \begin{sc}
    \begin{tabular}{l|c c|c}
    \toprule
    Model & FE & LD & MSE \\
    \midrule
    NP & & & 0.166 $\pm$ 0.002 \\
    ANP & & & 0.142 $\pm$ 0.002 \\
    \midrule
    NP+FE & \cmark & & 0.138 $\pm$ 0.002 \\
    NP+LD & & \cmark & 0.312 $\pm$ 0.002 \\
    \midrule
    FELD & \cmark & \cmark & \textbf{0.130 $\pm$ 0.002} \\
    \bottomrule
    \end{tabular}
    \end{sc}
    \end{scriptsize}
    \end{center}
\end{wraptable}
\paragraph{Gaussian Process} Following the basic neural processes \citep{garnelo2018conditional, garnelo2018neural, kim2019attentive}, we consider functions generated from GP with squared exponential kernel $k(x, x')=\sigma^2 \exp{\left(-0.5(x-x')^2/l^2\right)}$. The experimental result in Table~\ref{gp regression MSE} states that although ANP performs better than NP in terms of flexibility, the dominance no longer holds when NP is equipped with the flexible encoder. However, a degradation in performance is shown when using the linear decoder in NP. This is empirical evidence that NP strongly relies on the complexity of the decoder in regression, by which the model is prone to ignore the latent variables \citep{chen2016variational, zhao2017towards}. By exploiting the flexible encoder to obtain more informative latent variables by themselves such that the (shallow) linear decoder is just enough for prediction, FELD performs better than any other models with the (deep) conventional decoder. We find the Set Transformer is the perfect choice whose improvement can not be caught up by simply stacking MLPs. Moreover, it is noticeable that FELD outperforms NP+FE despite a decreased model capacity. 

\begin{wraptable}[17]{r}{0.46\textwidth}
    \caption{Accuracy on mini-ImageNet}
    \label{miniImageNet accuracy}
    \begin{center}
    \begin{scriptsize}
    \begin{sc}
    \begin{tabular}{l|c c}
    \toprule
    Model & 5-way 1-shot & 5-way 5-shot \\
    \midrule
    Matching Net & 43.40 $\pm$ 0.78\% & 51.09 $\pm$ 0.71\% \\
    Meta-LSTM & 43.44 $\pm$ 0.77\% & 60.60 $\pm$ 0.71\% \\
    MAML & 48.70 $\pm$ 1.84\% & 63.11 $\pm$ 0.92\% \\
    ProtoNet & 49.42 $\pm$ 0.78\% & 68.20 $\pm$ 0.66\% \\
    REPTILE & 49.97 $\pm$ 0.32\% & 65.99 $\pm$ 0.58\% \\
    Relation Net & 50.44 $\pm$ 0.82\% & 65.32 $\pm$ 0.70\% \\
    CAVIA & 51.82 $\pm$ 0.65\% & 65.85 $\pm$ 0.55\% \\
    VERSA & 53.40 $\pm$ 1.82\% & 67.37 $\pm$ 0.86\% \\
    TPN & 55.51 $\pm$ 0.86\% & 69.86 $\pm$ 0.65\% \\
    \midrule
    Meta-SGD & 54.24 $\pm$ 0.03\% & 70.86 $\pm$ 0.04\% \\
    SNAIL & 55.71 $\pm$ 0.99\% & 68.88 $\pm$ 0.92\% \\
    NP+LD & 57.30 $\pm$ 0.06\% & 75.10 $\pm$ 0.04\% \\ 
    TADAM & 58.50 $\pm$ 0.30\% & 76.70 $\pm$ 0.30\% \\
    LEO & 61.76 $\pm$ 0.08\% & 77.59 $\pm$ 0.12\% \\
    \midrule
    FELD & \textbf{62.77 $\pm$ 0.05\%} & \textbf{81.15 $\pm$ 0.03\%} \\
    \bottomrule
    \end{tabular}
    \end{sc}
    \end{scriptsize}
    \end{center}
\end{wraptable}

\paragraph{Mini-ImageNet, Tiered-ImageNet} Similar tendency can be observed in classification. We consider mini-ImageNet \citep{vinyals2016matching} and tiered-ImageNet \citep{ren2018meta}, which are frequently used large-scale datasets for few-shot image classification. 
For mini-ImageNet, we follow the split of \cite{ravi2016optimization}, which assigns 64 classes for the meta-train set, 16 classes for the meta-valid set, and 20 classes for the meta-test set. For tiered-ImageNet, 608 classes are first grouped into 34 higher-level nodes, divided into 20, 6, and 8 nodes to construct the meta-train set, meta-valid set, and meta-test set. 
We use the feature provided by \citep{rusu2018meta}, which is obtained by pre-training a deep residual network in a supervised manner as in \citep{gidaris2018dynamic, oreshkin2018tadam, qiao2018few}. However, unlike \citep{qiao2018few, rusu2018meta}, the meta-valid set is used for early stopping and hyperparameter search but not utilized to update the parameters. 

\begin{wraptable}[14]{r}{0.46\textwidth}
    \caption{Accuracy on tiered-ImageNet}
    \label{tieredImageNet accuracy}
    \begin{center}
    \begin{scriptsize}
    \begin{sc}
    \begin{tabular}{l|c c}
    \toprule
    Model & 5-way 1-shot & 5-way 5-shot \\
    \midrule
    MAML & 51.67 $\pm$ 1.81\% & 70.30 $\pm$ 0.08\% \\
    ProtoNet & 53.31 $\pm$ 0.89\% & 72.69 $\pm$ 0.74\% \\
    Relation Net & 54.48 $\pm$ 0.93\% & 71.32 $\pm$ 0.78\% \\
    Warp-MAML & 57.20 $\pm$ 0.90\% & 74.10 $\pm$ 0.70\% \\ 
    TPN & 57.41 $\pm$ 0.94\% & 71.55 $\pm$ 0.74\% \\
    \midrule
    Meta-SGD & 62.95 $\pm$ 0.03\% & 79.34 $\pm$ 0.06\% \\
    NP+LD & 63.36 $\pm$ 0.06\% & 80.50 $\pm$ 0.04\% \\
    LEO & 66.33 $\pm$ 0.05\% & 81.44 $\pm$ 0.09\% \\
    \midrule
    FELD & \textbf{66.87 $\pm$ 0.06\%} & \textbf{83.54 $\pm$ 0.04\%} \\
    \bottomrule
    \end{tabular}
    \end{sc}
    \end{scriptsize}
    \end{center}
\end{wraptable}
    
In Table~\ref{miniImageNet accuracy}, \ref{tieredImageNet accuracy}, accuracy on mini-ImageNet and tiered-ImageNet is reported. We collect the score of various baselines that use either convolutional networks or deep residual networks and do not exploit any data augmentation for a fair comparison. While NP performs no better than a random guess when following \citep{garnelo2018conditional}, NP+LD results in a comparable score to the recent models in gradient-based meta-learning, verifying the validity of the linear decoder in classification. FELD achieves even better performance than the state-of-the-art, which is remarkable in the sense that the attention modules in Set Transformers can not be fully utilized in low-\textit{shot} regime. 

\subsection{Heterogeneous dataset}
\label{heterogeneous dataset}

\paragraph{Sine $\&$ Polynomial} To verify the performance on the family of functions, we experiment on the toy 1D regression as in \citep{vuorio2018toward, yao2019hierarchically, yao2020automated}. In particular, we follow the exact setting of \citep{yao2019hierarchically} where each task is randomly chosen to be one of the following one-dimensional functions where the coefficients are uniformly sampled from the prefixed intervals summarized in Appendix C.1: (sine) $y = A_s sin(B_s x) + C_s$, (line) $y = A_l x + B_l$, (quad) $y = A_q x^2 + B_q x + C_q$, (cubic) $y = A_c x^3 + B_c x^2 + C_c x + D_c$. A small number of data points are given as context, requiring the model to appropriately interpolate and extrapolate in a highly variable prediction space. 

\begin{wraptable}[13]{r}{0.44\textwidth}
    \caption{MSE on Sine $\&$ Polynomial}
    \label{s/l/q/c regression MSE}
    \begin{center}
    \begin{scriptsize}
    \begin{sc}
    \begin{tabular}{l|c c}
    \toprule
    Model & 5-shot & 10-shot \\
    \midrule
    BMAML & 2.435 $\pm$ 0.130 & 0.967 $\pm$ 0.056 \\
    MAML & 2.205 $\pm$ 0.121 & 0.761 $\pm$ 0.068 \\
    META-SGD & 2.053 $\pm$ 0.117 & 0.836 $\pm$ 0.065 \\
    MT-NET & 2.016 $\pm$ 0.019 & 0.698 $\pm$ 0.054 \\
    MUMOMAML & 1.096 $\pm$ 0.085 & 0.256 $\pm$ 0.028 \\
    HSML & 0.856 $\pm$ 0.073 & 0.161 $\pm$ 0.021 \\
    NP & 0.514 $\pm$ 0.051 & 0.089 $\pm$ 0.015 \\
    ANP & 0.415 $\pm$ 0.046 & 0.058 $\pm$ 0.016 \\
    \midrule
    FELD & 0.118 $\pm$ 0.015 & 0.008 $\pm$ 0.002 \\ 
    MAHA & 0.077 $\pm$ 0.006 & 0.003 $\pm$ 0.001 \\
    MAHA* & \textbf{0.056 $\pm$ 0.003} & \textbf{0.002 $\pm$ 0.001} \\
    \bottomrule
    \end{tabular}
    \end{sc}
    \end{scriptsize}
    \end{center}
\end{wraptable}

In Table~\ref{s/l/q/c regression MSE}, MSE over 4000 tasks are presented with 95\% confidence interval. Generally, all the gradient-based meta-learning algorithms are outperformed by the neural processes, and a noticeable gain is again observed by solely exploiting the encoder-decoder pipeline, FELD. By adjusting FELD to MAHA by task clustering and MAHA to MAHA* by knowledge distillation, a monotonic improvement is observed.\footnote{We handle the overconfident nature of deep learning to better cope with the ambiguity by distilling an obtainable knowledge from $T$ to $C$. Please refer to Appendix B for a more detailed explanation.}

\begin{figure*}[t]
     \centering
     \includegraphics[width=\textwidth]{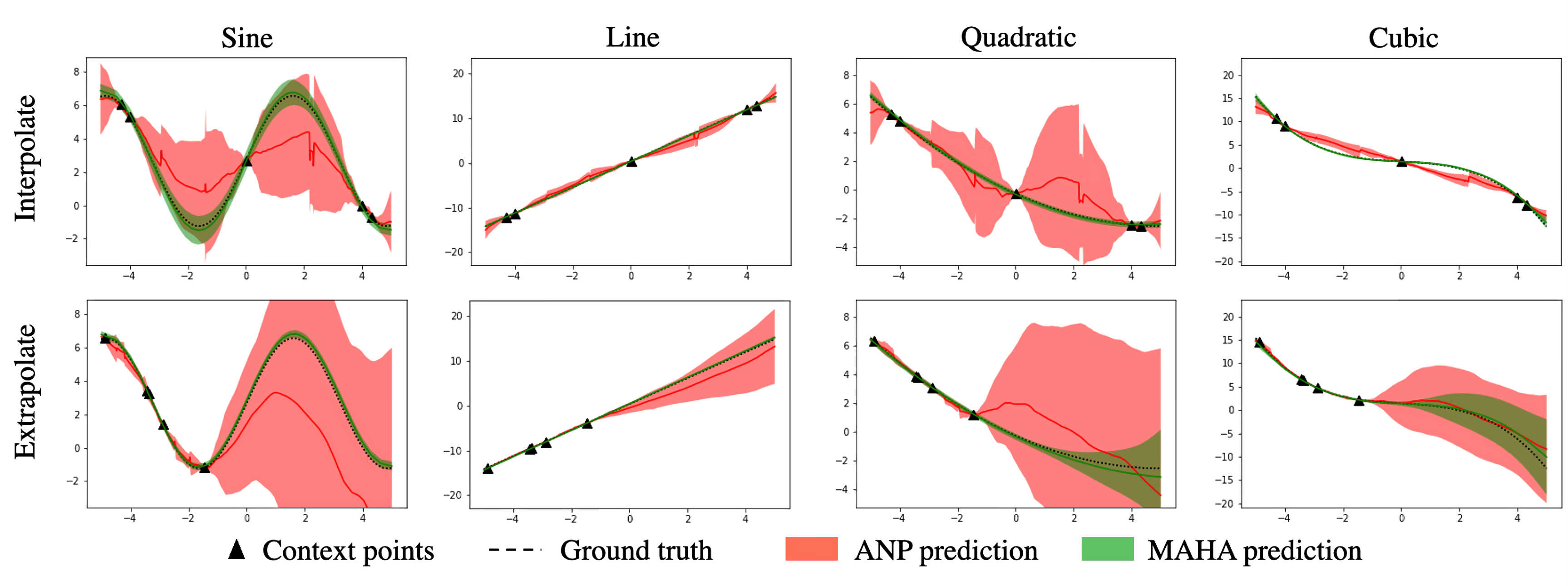}
     \caption{Qualitative comparison of ANP and MAHA on various function types. The context points are selected from 40\% of the entire domain for extrapolation.}
     \label{regression performance}
\end{figure*}

In Figure~\ref{regression performance}, we illustrate the interpolation and extrapolation of MAHA in comparison to ANP. As noted in Section~\ref{encoder-decoder pipeline}, the main interest of ANP is shown to fitting the context points, which poorly perform in predicting the target outputs whose corresponding inputs are located farther away from that of the context points. This tendency can be observed during interpolation and extrapolation, leading to a wiggly prediction with significant variance. By contrast, MAHA can correctly infer the functional shape, which can be confirmed through a consistently low variance.


\begin{wrapfigure}[18]{r}{0.345\textwidth}
    \centering
    \includegraphics[width=0.32\textwidth]{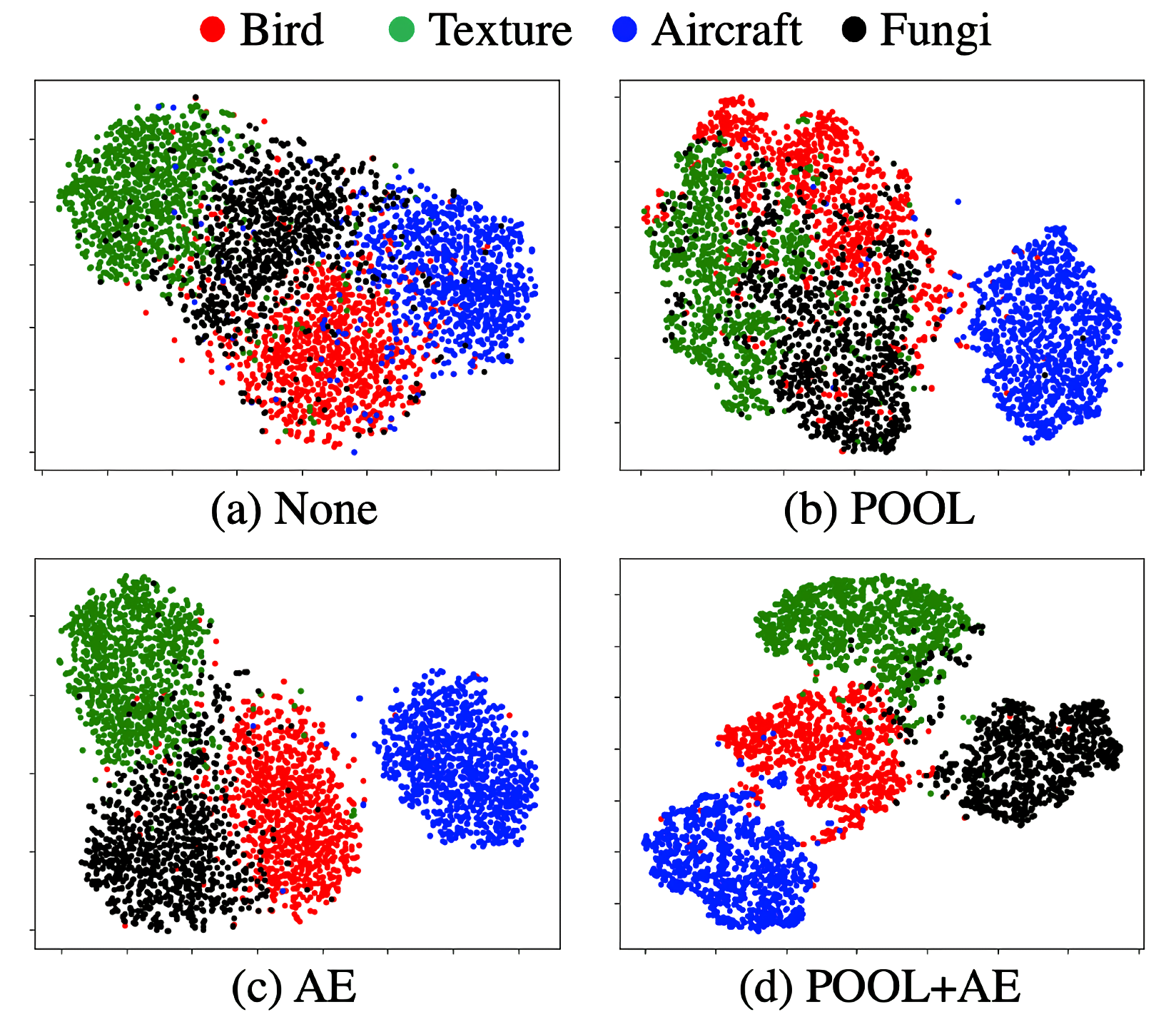}
    \begin{scriptsize}
    \begin{sc}
    \begin{center}
    \begin{tabular}{c c|c c}
    \toprule
    POOL & AE & 1-shot & 5-shot \\
    \midrule
    & & 0.8020 & 0.9957 \\
    \cmark & & 0.7455 & 0.9145 \\
    & \cmark & 0.9035 & 0.9930 \\
    \cmark & \cmark & \textbf{0.9560} & \textbf{0.9992} \\
    \bottomrule
    \end{tabular}
    \end{center}
    \end{sc}
    \end{scriptsize}
    \caption{t-SNE of $\mu_{\bar{z}}$ from $q(\bar{z}|C)$ and the estimated purity values}
    \label{t-SNE}
\end{wrapfigure}

\paragraph{Multi-dataset} Four distinct fine-grained image classification datasets are combined to construct the multi-dataset proposed in \citep{yao2019hierarchically}: (Bird) CUB-200-2011, (Texture) Describable Textures Dataset, (Aircraft) FGVC of Aircraft, and (Fungi) FGVCx-Fungi. Compared to a homogeneous setting, this is more challenging since overfitting to a particular dataset can critically harm the performance. For the feature extractor, we followed \citep{yao2019hierarchically} where 2-Conv blocks are used for task clustering, and 4-Conv blocks are used for prediction.

In Figure~\ref{t-SNE}, for 1-\textit{shot} setting, mean value of the variational distribution $q(\bar{z}|C)$ is visualized through t-SNE \citep{van2008visualizing}. Without external knowledge, such as the number of true clusters, the embeddings get interpretable when using both the dimension-wise pooling and the auto-encoding structure. The distinct datasets are no more clearly discriminated without either of them, which is quantitatively demonstrated by the estimated purity values in the bottom table. Note that the validity of the methodologies stands out particularly in low-\textit{shot} regime which implies the difficulty of task identification within ambiguity. 

The tendency can be observed by the performance measure presented in Table~\ref{multi-dataset accuracy}.
Compared to 1-\textit{shot} setting where a noticeable gain is occurred by task clustering, in 5-\textit{shot} setting, there is almost no difference between FELD and MAHA. This is because the models can clearly identify the tasks regardless of whether the pooling or the auto-encoding structure is used or not, demonstrated by the high purity values. Accordingly, the knowledge distillation, which is fundamentally devised to regularize the model within ambiguity appropriately, has shown a worthwhile improvement from MAHA to MAHA* particularly in 1-\textit{shot} setting. Eventually, MAHA (and MAHA*) beats all the previous works with a fairly large margin and achieves state-of-the-art performance.

\begin{table}[t]
    \caption{Accuracy on multi-dataset}
    \label{multi-dataset accuracy}
    \begin{center}
    \begin{scriptsize}
    \begin{sc}
    \begin{tabular}{c | l | c c c c | c}
    \toprule
    & Model & Bird & Texture & Aircraft & Fungi & Average\\
    \midrule
    \multirow{11}{*}{5-way 1-shot} 
    & MAML & 53.94 $\pm$ 1.45\% & 31.66 $\pm$ 1.31\% & 51.37 $\pm$ 1.38\% & 42.12 $\pm$ 1.36\% & 44.77\% \\
    & Meta-SGD & 55.58 $\pm$ 1.43\% & 32.38 $\pm$ 1.32\% & 52.99 $\pm$ 1.36\% & 41.74 $\pm$ 1.34\% & 45.67\% \\
    & MT-NET & 58.72 $\pm$ 1.43\% & 32.80 $\pm$ 1.35\% & 47.72 $\pm$ 1.46\% & 43.11 $\pm$ 1.42\% & 45.59\% \\
    & BMAML & 54.89 $\pm$ 1.48\% & 32.53 $\pm$ 1.33\% & 53.63 $\pm$ 1.37\% & 42.50 $\pm$ 1.33\% & 45.89\% \\
    & MUMOMAML & 56.82 $\pm$ 1.49\% & 33.81 $\pm$ 1.36\% & 53.14 $\pm$ 1.39\% & 42.22 $\pm$ 1.40\% & 46.50\% \\
    & HSML & 60.98 $\pm$ 1.50\% & 35.01 $\pm$ 1.36\% & 57.38 $\pm$ 1.40\% & 44.02 $\pm$ 1.39\% & 49.35\% \\
    & ARML & 62.33 $\pm$ 1.47\% & 35.65 $\pm$ 1.40\% & 58.56 $\pm$ 1.41\% & 44.82 $\pm$ 1.38\% & 50.34\% \\
    \cmidrule(lr){2-7}
    & FELD & 56.17 $\pm$ 0.64\% & 35.86 $\pm$ 0.41\% & 53.03 $\pm$ 0.58\% & 45.41 $\pm$ 0.58\% & 47.61\% \\ 
    & MAHA & 63.89 $\pm$ 0.34\% & 37.22 $\pm$ 0.23\% & 58.90 $\pm$ 0.44\% & 47.95 $\pm$ 0.34\% & 51.99\% \\
    & MAHA* & \textbf{64.45 $\pm$ 0.36\%} & \textbf{37.83 $\pm$ 0.23\%} & \textbf{59.18 $\pm$ 0.43\%} & \textbf{48.33 $\pm$ 0.33\%} & \textbf{52.41\%} \\
    \midrule
    \multirow{11}{*}{5-way 5-shot} 
    & MAML & 68.52 $\pm$ 0.79\% & 44.56 $\pm$ 0.68\% & 66.18 $\pm$ 0.71\% & 51.85 $\pm$ 0.85\% & 57.78\% \\
    & Meta-SGD & 67.87 $\pm$ 0.74\% & 45.49 $\pm$ 0.68\% & 66.84 $\pm$ 0.70\% & 52.51 $\pm$ 0.81\% & 58.18\% \\
    & MT-NET & 69.22 $\pm$ 0.75\% & 46.57 $\pm$ 0.70\% & 63.03 $\pm$ 0.69\% & 53.49 $\pm$ 0.83\% & 58.08\% \\
    & BMAML & 69.01 $\pm$ 0.74\% & 46.06 $\pm$ 0.69\% & 65.74 $\pm$ 0.67\% & 52.43 $\pm$ 0.84\% & 58.31\% \\
    & MUMOMAML & 70.49 $\pm$ 0.76\% & 45.89 $\pm$ 0.69\% & 67.31 $\pm$ 0.68\% & 53.96 $\pm$ 0.82\% & 59.41\% \\
    & HSML & 71.68 $\pm$ 0.73\% & 48.08 $\pm$ 0.69\% & 73.49 $\pm$ 0.68\% & 56.32 $\pm$ 0.80\% & 62.39\% \\
    & ARML & 73.34 $\pm$ 0.70\% & 49.67 $\pm$ 0.67\% & 74.88 $\pm$ 0.64\% & 57.55 $\pm$ 0.82\% & 63.86\% \\
    \cmidrule(lr){2-7}
    & FELD & \textbf{77.63 $\pm$ 0.46\%} & \textbf{55.80 $\pm$ 0.38\%} & 75.88 $\pm$ 0.41\% & 63.68 $\pm$ 0.50\% &  68.24\% \\
    & MAHA & 75.04 $\pm$ 0.26\% & 54.39 $\pm$ 0.21\% & \textbf{79.98 $\pm$ 0.20\%} & \textbf{65.09 $\pm$ 0.25\%} & \textbf{68.62\%} \\
    & MAHA* & 75.82 $\pm$ 0.26\% & 54.28 $\pm$ 0.22\% & \textbf{79.91 $\pm$ 0.19\%} & \textbf{65.18 $\pm$ 0.25\%} & \textbf{68.79\%} \\
    \bottomrule
    \end{tabular}
    \end{sc}
    \end{scriptsize}
    \end{center}
\end{table}



\section{Conclusion}
\label{conclusion}

This paper proposes a new meta-learning framework, MAHA, that performs robustly amidst heterogeneity and ambiguity. We aim to disentangle the stochastic representation by the dimension-wise pooling and the auto-encoding structure based on the newly devised encoder-decoder pipeline to better leverage the latent variables. With the multi-step training process, comprehensive experiments are conducted on regression and classification. In the end, we argue that the proposed model captures the task identity with lower variance, leading to a noticeable improvement in performance. The potential limitation of MAHA would be the additional computational cost from the flexible encoder composed of multiple attention modules. However, by orthogonally applying to the existing work, the compatibility and the necessity are empirically verified. An interesting future work would be to apply our model to reinforcement learning. In particular, training a policy directly from well-clustered representations for sample-efficient exploration seems promising in an environment with sparse rewards.

\section*{Broader Impact}
\label{impact}

When training meta-learning models, there comes a customization process based on the problem at hand. If not using the benchmark datasets that frequently appear in academia, it becomes unclear to which extent the distinct datasets should be combined, expecting the model to be versatile on every possible task generation. MAHA, in this respect, can guide for a human to analyze and cluster the available data into separate clusters. Moreover, MAHA
mainly benefits future AI industries where the limited communication between the decentralized servers is available as it can infer the global context even with a small amount of information. As a result, we do not expect any negative societal impacts, but we believe that MAHA possesses many implications in more realistic scenarios.

\newpage

{
\small
\bibliography{neurips_2021}
\bibliographystyle{plainnat}
}

\end{document}